\pdfoutput=1
\documentclass[11pt]{article}

\usepackage{acl}

\usepackage{times}
\usepackage{latexsym}
\usepackage[T1]{fontenc}
\usepackage[utf8]{inputenc}

\usepackage{microtype}

\usepackage{inconsolata}

\usepackage{amsmath}
\usepackage{amssymb}
\usepackage{graphicx}
\usepackage{array,multirow}

\title{Adjusting Interpretable Dimensions in Embedding Space \\  with Human Judgments}

\author{Katrin Erk \\
  University of Texas at Austin \\
  \texttt{katrin.erk@utexas.edu} \\\And
  Marianna Apidianaki \\
  University of Pennsylvania \\
  \texttt{marapi@seas.upenn.edu} \\}

\begin{document}
\maketitle
\begin{abstract}
Embedding spaces contain interpretable dimensions indicating gender, formality in style, or even object properties. This has been observed multiple times. Such interpretable dimensions are becoming valuable tools in different areas of study, from social science to neuroscience. The standard way to compute these dimensions
uses contrasting seed words and computes  difference vectors over
them. This is simple but does not always work well. 
We combine seed-based vectors with guidance from 
human ratings of where words fall along a specific dimension, and evaluate on predicting both object properties like size and danger, and the stylistic
properties of formality and complexity. We obtain
interpretable dimensions with markedly better performance especially
in cases where seed-based dimensions do not work well. 
\end{abstract}

\section{Introduction}

Properties are commonly used in linguistics \citep{katz:structure,Jackendoff1990-JACSS,Wierzbicka1996-WIESPA} as well as in psychology \citep{Murphy:2002} for representing word meanings and concepts. Those same properties are discoverable as \textit{interpretable dimensions} in word embedding space, and can be used to explore the
patterns and regularities encoded by Large Language Models (LLMs) \citep{mikolov-etal-2013-linguistic,Bolukbasi-NIPS2016}. 
Because LLMs are trained on texts from many different authors, we can view them as a compact repository of human utterances.
This makes them an interesting resource for studying linguistic phenomena, analyzing social contexts of words, or as a stand-in for conceptual knowledge for interpreting brain voxels. Interpretable dimensions provide an attractive and simple way to access this resource~\citep{FedorenkoDimensions,kozlowskiGeometryCultureAnalyzing2019,gari-soler-apidianaki-2020-bert,lucy-etal-2022-discovering}.
Compared to probing \citep{tenney2018what,conneau-etal-2018-cram}, interpretable dimensions allow for a direct exploration of LLM embedding space, without external classifiers.

 \begin{figure}[tb]
  \centering
  \begin{tabular}{@{}ll}
    \includegraphics[width=9.5em]{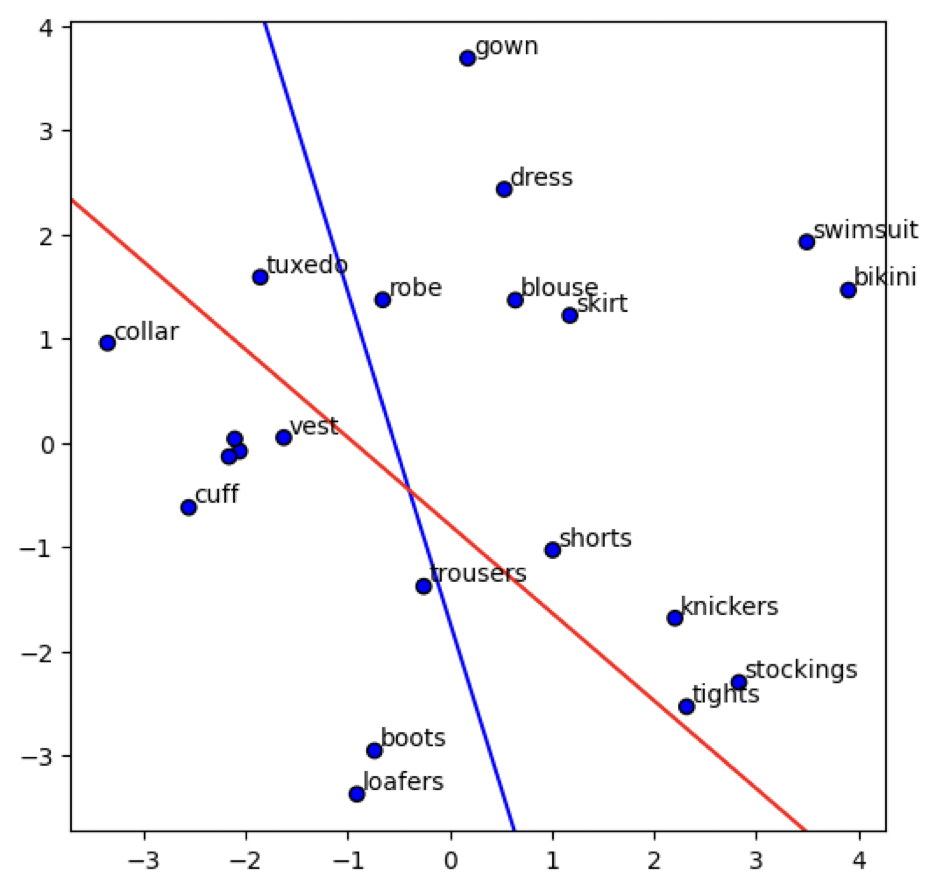} & \includegraphics[width=9.5em]{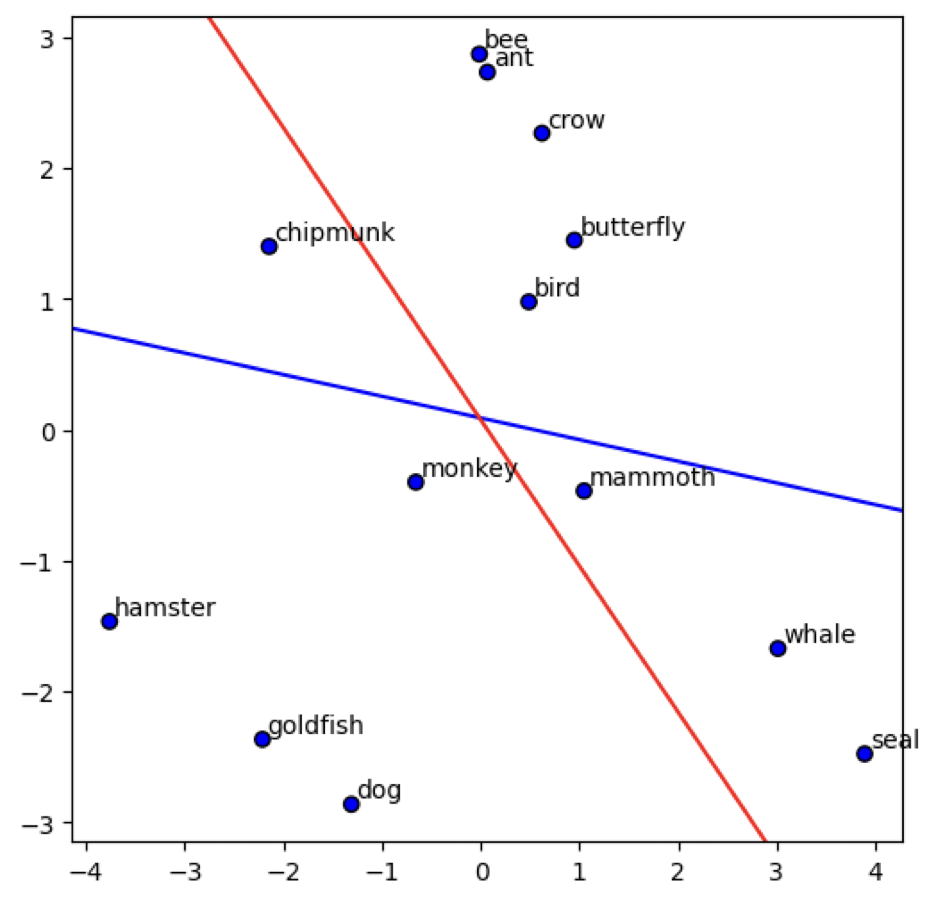}
  \end{tabular}
  \caption{Interpretable dimensions for two object categories and features from \citet{FedorenkoDimensions}: clothes by wealth (left), animals by size (right). PCA projection of embeddings, with seed-based (blue) and \textsc{fit+s} (red) dimensions.}
  \label{fig:dims}
\end{figure}

The most common way to obtain  interpretable dimensions is to  specify some seed pairs of antonyms, and take the average over their vector differences. 
But it is unclear what makes good seed pairs, or even how to test whether a particular property corresponds to a discernible dimension in embedding space. \citet{antoniak-mimno-2021-bad} and \citet{lucy-etal-2022-discovering} express concerns about the quality of commonly used hand-curated seed lexicons and propose metrics for evaluating seeds.

In this paper, we take a different approach to addressing the problem of ``bad seeds''~\citep{antoniak-mimno-2021-bad}: We propose a method to augment seed-based interpretable dimensions with additional guidance from human ratings, and we show that this augmentation is particularly impactful when the original seed-based dimensions are problematic.\footnote{Our code and data are available at \url{https://github.com/mariannaapi/interpretable-dimensions}.} Figure~\ref{fig:dims} shows words for clothes with dimensions for wealth, and animal names with dimensions for size, blue for seed-based dimensions and red for our new fitted dimensions. The fitted dimensions correct overly high wealth estimates for \textit{tights} and \textit{stockings}, and exaggerated size estimates for \textit{bee} and \textit{ant}.

While interpretable dimensions are useful both to social science and to cognitive science, there is an important difference between the fields: 
In social science, crowd-sourced datasets cannot be trusted in absolute terms, because annotators may have social biases of their own; this is  the scenario that \citet{antoniak-mimno-2021-bad} address. In cognitive science however, experimental data from human participants is central (though the method used to solicit data can have an influence on the outcome). We work with data from cognitive science  here, so we can use human ratings to improve seed dimensions.

Our new method draws inspiration from a completely separate strand of research on interpretable dimensions, in the context of knowledge graph embeddings \citep{derracInducingSemanticRelations2015,JameelSchockaert,bouraoui_et_al:OASIcs.AIB.2022.3}. There, interpretable dimensions are learned using labeled training data. In the current paper, we use a similar learning strategy, and apply it to a combination of seed-based dimensions and labeled training data. We apply this technique to predict human ratings on object properties and stylistic aspects of words, and find that it improves performance particularly in cases where seed-based dimensions underperform, and that in contrast to seed-based dimensions it is able to make predictions at the same scale as the original ratings.

A larger issue is: When can we trust that an interpretable dimension shows us what the LLM truly ``knows'' about the property in question, that we are not misled by noise in our tool? This same worry  also arises for probing classifiers~\citep{hewitt-liang-2019-designing,belinkov-2022-probing}.\footnote{There is a related question of whether information encoded in the model is also relevant for downstream performance \citep{ravichander-etal-2021-probing,lyu2024faithful}. This question is not so central for linguists or cognitive scientists interested in knowledge reflected in an LLM.  What is central for them is that the dimension really encodes the property of interest.} We take one step towards addressing this issue: By combining two sources of information, seeds and human annotation, we hope to reduce noise present in either source.

\section{Related Work}

Interpretable dimensions in word embedding space have first been observed in NLP  
\citep{Bolukbasi-NIPS2016}, and the idea was then taken up in neuroscience and social science. \citet{FedorenkoDimensions} discover dimensions for objects' scalar properties (e.g., {\sc danger}, {\sc size}, {\sc speed}, {\sc wealth}). 
\citet{kozlowskiGeometryCultureAnalyzing2019} identify dimensions including {\sc affluence}, {\sc gender}, {\sc race} and {\sc morality}, and show that concepts such as sports (e.g., \textit{golf, boxing}) and music genres (e.g., \textit{opera, rap, jazz}) are ordered along these axes in ways that match cultural stereotypes.  
\citet{Gargetal:2018} explore ethnic stereotypes, and \citet{StoltzandTaylor:2021} go as far as to propose a cultural cartography with word embeddings. \citet{an-etal-2018-semaxis} use a large number of dimensions to characterize sentiment, which \citet{Kwak} apply to whole documents. Interpretable dimensions have also been used to represent linguistic notions, such as complexity and scalar adjective intensity \citep{gari-soler-apidianaki-2020-bert,gari-soler-apidianaki-2021-scalar,lyu-apidianaki-ccb:2023}. In our work, we explore dimensions addressing object properties in the \citet{FedorenkoDimensions} datasets, and the abstract  notions of formality and complexity. 

In all these studies, dimensions are discovered using the seed-based methodology, where a few seed pairs of antonyms 
are specified and the dimension is computed as the average over vector differences for these pairs. 
This method is simpler than alternative representation approaches 
(e.g., the multi-task learning framework of \citet{allaway-mckeown-2021-unified}).

Seed pair selection has until now been ad hoc; but some choices, such as the selected word pairs, their number and order, and the way they are combined, have a strong impact on the quality of the derived dimension. 
\citet{antoniak-mimno-2021-bad} address the ``bad seeds'' problem by  measuring the \textit{coherence} of each seed set  pairing 
after mapping to the bias subspace: When all words in the vocabulary are projected  onto the subspace, the two seed sets must be drawn as far apart as possible. 
\citet{lucy-etal-2022-discovering} propose to measure the semantic axis' \textit{self-consistency} using a leave-one-out approach, where each seed is compared to an axis constructed from the remaining seeds. 
A good seed, when left out, 
should be closer to the pole it belongs to. 

In our approach, we do not test for seed quality. Instead, we use human ratings to improve on seed-based dimensions. 
Our approach is inspired by work on knowledge graph embeddings \citep{derracInducingSemanticRelations2015,JameelSchockaert,Bouraoui2020}. 
Drawing on the conceptual spaces of \citet{Gardenfors:2014} for intuition, \citet{JameelSchockaert} learn embeddings of knowledge graph nodes that include interpretable dimensions for properties. 
Like us, they learn interpretable dimensions using labeled training data. Our objective function is an adaptation of their objective function, but still different as they also learn the space while we have a fixed space.

For constructing interpretable dimensions, most previous work used static embeddings (GloVe \citep{pennington2014glove} and word2vec \citep{word2vec}). Recent work extends the methodology to contextualized representations
\citep{gari-soler-apidianaki-2020-bert,lucy-etal-2022-discovering}. We experiment with both kinds of embeddings.

\section{Methods}

\subsection{Models}
\label{sec:models}

\paragraph{Seed-based dimensions (\textsc{seed} model).} 
The seed-based method is the most commonly used for computing interpretable dimensions~\citep{Bolukbasi-NIPS2016,kozlowskiGeometryCultureAnalyzing2019,devandphilips2019,gari-soler-apidianaki-2021-scalar,FedorenkoDimensions}. 
A group of \emph{seed words} are chosen which represent opposite ends of the dimension. For the \textsc{danger} dimension in \citet{FedorenkoDimensions}, for example, the seeds are \{\textit{safe, harmless, calm}\} for the positive side and \{\textit{dangerous, deadly, threatening}\} for the negative side. For each pair of a positive and negative seed word $p, n$ with vectors $\vec p, \vec n$, the difference vector $\vec p - \vec n$ is computed; this is a first estimate of the interpretable dimension, but the vectors can differ substantially across seed pairs. To obtain a more stable estimate, the vector for the interpretable dimension is then computed as the average of the difference vectors from individual seed pairs. The rating of any word $a$ on the property $d$ with interpretable dimension $\vec d$ -- in our example from above, {\sc danger} -- is then predicted as the scalar projection onto the dimension:
\[ ||\text{proj}_{\vec a}(\vec d)|| = \frac{\vec a \cdot \vec d}{||\vec d||}\]

\paragraph{Fitted dimensions (\textsc{fit} model).} Whenever we have gold ratings on some dimension, like human judgments on degrees of danger of different animals~\citep{FedorenkoDimensions} or gold ratings for complexity~\citep{lyu-apidianaki-ccb:2023}, we can estimate a direction in embedding space that best matches the gold ratings. We adapted an idea from \citet{JameelSchockaert}, who learn an embedding space for knowledge graph nodes in such a way that properties of the nodes correspond to dimensions in space. But rather than learning a new space, we need to use an existing space spanned by static or contextualized embeddings, because it is these spaces, and the patterns in human language use that they encode, that we want to analyze.

We proceed as follows. Let $W = \langle w_1, \ldots w_n\rangle$ be an annotated dataset of $n$ words with real-valued gold ratings $\hat Y = \langle \hat y_1, \ldots, \hat y_n\rangle$ for some feature $f$. Let $\vec w_i$ be the embedding of word $w_i$. For the dimension $\vec f$ to be computed for feature $f$ in that same embedding space, we stipulate that the scalar projection of $\vec{w_i}$ onto $\vec{f}$ be proportional to the gold rating $\hat y_i$. 
For example, say the gold rating (average human rating) of {\it dolphin} on the {\sc danger} scale (on a scale of 1-5) is 2.1, and the gold rating of {\it tiger} is 4.9. Then we want the length of the projection $\text{proj}_{\vec{\text{dolphin}}}(\vec{\text{{\sc danger}}})$ to be proportional to $c_{\text{{\sc danger}}} \cdot 2.1 + b_{\text{{\sc danger}}}$, 
and the length of the projection $\text{proj}_{\vec{\text{tiger}}}(\vec{\text{{\sc danger}}})$ to be proportional to $c_{\text{{\sc danger}}} \cdot 4.9 + b_{\text{{\sc danger}}}$, for some weight and bias constants $c_{\text{{\sc danger}}}, b_{\text{{\sc danger}}} \in \mathbb{R}$. So in general, we would like to have 
\[ \frac{\vec{w_i} \cdot \vec{f}}{||\vec f||} = c_f \:\hat y_i + b_f\]
We turn this into a loss function for computing a \emph{fitted dimension} $f$, dropping the denominator $||\vec f||$:
\[J_f = \sum_{w_i} \big( \vec{w_i} \cdot \vec{f} - c_f \:\hat y_i - b_f\big)^2\]

\paragraph{Fitted dimensions with seed words (\textsc{fit+sw} model).} We also test whether fitted dimensions can be guided by the seed words used to make seed-based dimensions. The first method follows the intuition of \citet{antoniak-mimno-2021-bad} that the scalar projections of seed words should sit ``far out'' on an interpretable dimension, further than other words. The \textsc{fit+sw} model simply extends the collection $W$ of rated words by the seed words. We make synthetic gold ratings for the seedwords, giving them extreme ratings: $\max(\hat Y) + o$ for positive seed words, and $\min(\hat Y) - o$ for negative seedwords, for an offset $o$ that is a hyperparameter. We optionally add a small amount of random jitter (sampled from the interval $[0.001, 0.005]$) so that the seed words don't all have the same rating.

\paragraph{Fitted dimensions with seed dimensions (\textsc{fit+sd} model).} Our second way of extending fitted dimensions with seed word information is built on the idea that seed-based dimensions and human ratings both provide useful information for fitting an interpretable dimension, and that they should be combined. So we use an overall loss function of 
\[J = \alpha J_f + (1-\alpha) J_d(D)\]
where $J_f$ is the loss function from above, and $\alpha$ is a hyperparameter. $J_d(D)$ is a loss that measures distance of the fitted dimension $\vec f$ from a set $D$ of seed-based dimensions, defined as 
\[J_d(D) = \sum_{d \in D} 1 - cosine(\vec d, \vec f)\]

\paragraph{Fitted dimensions with seeds as words and dimensions (\textsc{fit+s} model).} Our final model uses seeds both as seed words, as in \textsc{fit+sw}, and as seed dimensions, as in \textsc{fit+sd}.

\paragraph{Baselines.} We compare our methods to a  baseline which ranks words by frequency (FREQ). Frequency has been a strong baseline for complexity and formality in previous work, given that rare words tend to be more complex than frequently used words \citep{brooke-etal-2010-automatic}. We use log-transformed frequency counts in the Google N-gram corpus \citep{brants2006web}. 
We also use a random baseline, which assigns to each word a randomly selected score in the range [-3,~3].\footnote{This range was chosen because all gold ratings in our study are z-scores.}

\subsection{Evaluation metrics}
\label{sec:metrics}
In contrast to interpretable dimensions computed from seed words, the \textsc{fit} models use training data: words with human ratings for  
the property in question. When we evaluate these models, we use up some of the data for training, leaving less for testing. To mitigate the issue, we do cross-validation, and we focus on evaluation metrics that work well with smaller test sets. We do not use the correlation metrics used in \citet{gari-soler-apidianaki-2020-bert} and \citet{FedorenkoDimensions}, as significance tests become unreliable with small datasets. Instead, we use a variant of pairwise rank accuracy, a metric used in \citet{gari-soler-apidianaki-2020-bert}, \citet{cocos-etal-2018-learning},  and  \citet{FedorenkoDimensions}. 

Pairwise rank accuracy measures the percentage of pairs of words whose ordering in the gold ratings is the same as in the model predictions. We define a new variant which we call \textbf{extended pairwise rank accuracy, r$^+$-acc}, which measures pairwise rank  accuracy among words in the test set, and additionally pairwise rank accuracy between each test word and each training word. 
For example, if \textit{tiger} and \textit{butterfly} are in the training set for \textsc{danger}, and \textit{cat} is in the test, we check whether the score assigned to \textit{cat} ranks it after \textit{tiger} and before \textit{butterfly}. 
This metric gives us more evidence on the quality of predictions than pairwise rank accuracy on its own because it includes more comparisons, thus making the metric less sparse.
Let $<_g, <_m$ be two complete orderings of the words in $W$, the gold and model orderings, respectively. For words $w, w'$ in $W$, we define an auxiliary function \textit{rm} for ``rank match'': 
\[rm^{}_{<_g, <_m}(w, w') = \left\{\begin{array}{l@{}l}
1 & \text{ iff } (w <_g w' \wedge w <_m w') \\
& ~\vee (w >_g w' \wedge w >_m w')\\
0 & \text{ else}
\end{array}\right.\]

Then standard pairwise rank accuracy on $W = \langle w_1, \ldots, w_n\rangle$ is defined as
\[\begin{array}{@{}l}
\text{r-acc}_W(<_g, <_m) = \frac{1}{n (n -1)}\\
~~~~~~~~~~\sum_{1 \le i < j \le n} rm_{<_g, <_m}(w_i, w_j)\end{array}\]
Now assume that $T = \{k_1, \ldots, k_\ell\}$, with $k_j\in \{1, \ldots, n\}$ for all $j$, is the set of test word indices among the indices of $W$. Assume both orderings, $<_g$ and $<_m$, are defined on all of $W$. Then our new extended pairwise rank accuracy is 
\[\begin{array}{@{}l}
\text{r$^+$-acc}_W(<_g, <_m) = \frac{1}{\ell(\ell -1) + \ell(n-\ell)} \\
~~~~~~~~~~~~\sum_{1 \le i < j \le \ell} rm_{<_g, <_m}(w_{k_i}, w_{k_j}) +\\
~~~~~~~~~~~~\sum_{i \in \{1, \ldots, \ell\}, j \not\in T} rm_{<_g, <_m}(w_{k_i}, w_j)\end{array}\]
The first half of this formula measures pairwise rank accuracy among members of the test set; the second half measures rank accuracy of test words with respect to training words.

Pairwise rank accuracy and extended pairwise rank accuracy 
are similar to correlation metrics in that they measure to what extent gold and model-based rankings agree. And in fact all three metrics are highly correlated: We tested correlation between the three metrics on seed-based dimensions for the  \citet{FedorenkoDimensions} data and obtained highly significant correlations ($p\ll 0.0001, r = 0.972$ for pairwise  rank accuracy, $r = 0.971$ for 
extended pairwise rank accuracy).\footnote{To measure extended pairwise rank accuracy, 
the data was split into training and test folds in 5-fold cross-validation, and r$^+$-acc scores were averaged over folds. It is not surprising that the values are almost the same, as in this case extended pairwise rank accuracy 
is almost the same as standard pairwise rank accuracy except that 
the latter omits some pairwise comparisons, namely between training data points.}

As a second evaluation metric, we test how far off from the gold ratings each individual predicted rating is. We use the \textbf{mean square error (MSE)} of predicted ratings compared to gold ratings. We can do this because all \textsc{fit} models learn to predict ratings on the same scale as the gold ratings. In order to apply the same evaluation to the \textsc{seed} model and the baselines, we simply use linear regression to map model predictions to ratings on the same scale as the gold ratings. Linear regression models are fit on the training portion of each data set, so that test words remain truly unseen.

\subsection{Data and Vectors}
We use the ratings collected by \citet{FedorenkoDimensions} which describe properties of objects in nine categories:\footnote{The data is available on the
Open Science Framework at \url{https://osf.io/5r2sz/}. No license information is given in the repository.} 
animals, clothing, professions, weather phenomena, sports, mythological creatures, world cities, states of the United States, and first names.\footnote{Most categories consist of 50 items.} Each category is matched with a subset of these semantic features: age, arousal, cost, danger, gender, intelligence, location (indoors vs. outdoors), partisanship (liberal vs. conservative), religiosity, size, speed, temperature, valence, volume, wealth, weight, and wetness. 
For style, we use datasets released by \citet{pavlick-nenkova-2015-inducing} which contain words and phrases with human ratings of 
formality and complexity. 
For each dimension, we sample words\footnote{We sample words with more than three characters to exclude pronouns, articles, numerals, and multiword phrases.} 
with high annotation confidence (i.e. where annotators agreed about the word being complex or formal):
We calculate the mean standard deviation for words in our sample, and keep words where deviation between human scores is lower than that mean. 
The filtered datasets contain 1,160 words for complexity, and 1,274 words for formality. 

We extract seed words from two other datasets released by \citet{pavlick-nenkova-2015-inducing} which contain pairwise paraphrase judgments of formality and complexity.\footnote{The data is available at {\url{https://cs.brown.edu/people/epavlick/data.html}}, under ``Style Lexicons: Human and automatic scores of formality and complexity for words, phrases, and sentences''. No license for the data is given.} Annotations reflect which phrase  in a pair (e.g., \textit{letter}-\textit{communication}, \textit{largely}-\textit{extensively}) is more complex or formal than the other. We collect five pairs of words for each style dimension for which inter-rater agreement is high. For complexity, we obtain the seed pairs 
 {\it work - employment, further - subsequently, strong - powerful, train - railway, shown - indicated}, where the first member of each pair is the negative seed (the simpler word). 
For formality, we used {\it winner - recipient, terrible - disastrous, membership - affiliation, highest - paramount, test - verify}, where again the first member of each pair is the negative seed (the less formal word).

Following \citet{FedorenkoDimensions}, we averaged over human subject ratings for each datapoint, then normalized ratings to z-scores separately for each pair of a category and property.\footnote{We would expect the data to show subjective differences between annotators, and in the future we would like to model the subjective ratings directly, following \citet{plank-etal-2014-linguistically}.} For formality and complexity, ratings were also converted to z-scores. 

As embeddings, we use the same representations as \citet{FedorenkoDimensions}, pre-trained GLoVE embeddings trained on the Common Crawl (42b tokens), 300 dimensions, uncased \citep{pennington2014glove}. 
We also use contextualized representations from the BERT ({\tt bert-large-uncased}) and RoBERTa ({\tt roberta-large}) 
models \citep{devlin2019bert,liu2019roberta} with sentences from UkWac  \citep{baroni2009wacky}.\footnote{Details about sentence selection are given in the Appendix.} 
For each word instance, we average its contextualized representations from the top 4 layers of the model.\footnote{ Averaging across layer subsets is generally better than averaging across all layers or selecting a single layer \citep{vulic-etal-2020-probing}.} 
If the word is split into multiple wordpieces during tokenization, we average the representations of its pieces  in order to obtain a single type-level representation for each word, as is common practice in semantic probing studies \citep{bommasani-etal-2020-interpreting,vulic-etal-2020-probing,GariSolerandApidianaki:TACL2021}. 
The final representation for a word is the average of its representations from the available sentences. 
Aggregating representations across multiple contexts is the most common approach for creating word type-level embeddings from contextualized representations    
which serves to null out, to some extent, the impact of specific contexts \citep{10.1162/coli_a_00474}. 
It is possible to use more sophisticated ways for sentence selection, such as language modeling criteria \citep{gari-soler-apidianaki-2020-bert} and exclusion of contexts where antonyms could occur \citep{lucy-etal-2022-discovering}. However applying such sophisticated context selection methods is not always better than random selection, which might be due to the skewed distribution of word senses  and the stronger presence of the most frequent sense of a word in randomly selected sentences \citep{Kilgarriff2004,mccarthy-etal-2004-finding}.

\section{Results and discussion}

In this section we evaluate the different interpretable dimension models from  Section \
\ref{sec:models} on the tasks of predicting human ratings of 
object properties \citep{FedorenkoDimensions}, and human ratings of 
the complexity and formality of words
\citep{pavlick-nenkova-2015-inducing}.

\paragraph{Experimental setup.} To make the most of the limited
available data, all models were tested in
5-fold cross-validation. In addition, all models that involve
randomness (all except \textsc{seed}) were re-run three times with
different random seeds. 
For Grand et al. object features, we first
compute mean \textit{r$^+$-acc} and median MSE for each
category/property pair 
(averaging over cross-validation runs and random seeds), then we
report averages over those. For formality and complexity we report
overall mean \textit{r$^+$-acc} and median MSE.\footnote{We use median MSE
because outliers make the mean uninformatively high.} 
Note that because we split the data into training and test using
cross-validation, the numbers reported in this paper are not
comparable with those reported in earlier papers on the same dataset.
We do however compute \textsc{seed} dimensions with the same
cross-validation setup as the \textsc{fit}-based dimensions, so that
the numbers that we report are comparable to each other.

To set hyperparameters, we sample 6 category/property pairs from the Grand et al.\  data as development
set. Hyperparameters were optimized once per embedding space; there
was no separate hyperparameter optimization for the formality and
complexity data.\footnote{Human ratings on all datasets were on the
  same scales as we normalized them all to z-scores.}  Overall, we
find that low values of $\alpha$ work well, and that it is beneficial
to input a single averaged seed dimension to \textsc{fit+sd} and
\textsc{fit+s}, rather than individual seed dimensions. The choice of offset
and jitter does not matter. More details
on hyperparameters and the development set can be found in the Appendix. Results reported
below for Grand et al.\ data are for all category/property pairs not in the
development set.

 \begin{table*}[tb]
    \centering
    \small
    \begin{tabular}{ll|*{5}{r@{$\:$}r}|l}  
      && \multicolumn{2}{c}{\textsc{seed}} & \multicolumn{2}{c}{\textsc{fit}} & \multicolumn{2}{c}{\textsc{fit+sw}}  & \multicolumn{2}{c}{\textsc{fit+sd}}  & \multicolumn{2}{c}{\textsc{fit+s}} &  \\ \hline 
      \multirow{2}{*}{GLoVE} & r$^+$-acc & 0.64 & (0.1) & 0.54 & (0.03) & 0.53 & (0.03) & 0.65 & (0.1) & \textbf{0.80} & (0.06) &  \multirow{5}{*}{ \parbox{1.8cm}{\vspace{2mm} {\sc freq} \\ r$^+$-acc: 0.58 \vspace{2mm} \\ {\sc rand} \\ r$^+$-acc: 0.49}} 
      \\ 
& MSE & $>1000$ &($>1000$) & 113.2 &(111.7) &  177.1& (125.4) & 89.6&(199.6) &  \textbf{0.7} & (0.36) & \\\cline{1-12}

\multirow{2}{*}{BERT} & r$^+$-acc & 0.64 &(0.1) & 0.51& (0.03) & 0.52 &(0.03) & 0.66 &(0.10) & \textbf{0.71}& (0.04) & \\
& MSE & $>1000$& ($>1000$) & 417.4& (271.7) & 597.4 &(525.6) & 115.0 &(437.4) &  \textbf{2.0}& (0.6) & \\\cline{1-12}

\multirow{2}{*}{RoBERTa} & r$^+$-acc & 0.57& (0.08) & 0.51& (0.03) & 0.51& (0.03) & 0.60& (0.1) & \textbf{0.69} &(0.04) & \\
& MSE & $>1000$ &($>1000$) & 392.5& (291.3) & 458.0& (284.3) & 125.2
            & (270.5) &    \textbf{1.9} &   (0.6) & \\ 
    \end{tabular}
    \caption{
    Results on {\bf object properties}: Extended rank accuracy (abbreviated r$^+$-acc) and Mean Squared Error (MSE), averaged over category/property pairs. In brackets:
      Standard error. Shown for 3 embedding spaces. Bolded: best
      performance for each embedding.}
    \label{tab:grand}
  \end{table*}

\begin{table*}[tb]
  \centering
  \small
    \begin{tabular}{p{1.1cm}p{8.5mm}|p{5.8mm}p{6mm}p{1cm}p{6.4mm}p{6.4mm}|p{8.6mm}||p{5.8mm}p{6mm}p{6.7mm}p{6mm}p{6.4mm}|p{7mm}}
      && \multicolumn{6}{c}{Complexity} & \multicolumn{6}{c}{Formality}\\ \cline{3-14}
      && \parbox{6mm \centering \textsc{seed}} & \parbox{6mm \centering \textsc{fit}} & \parbox{1cm \centering\textsc{fit+sw}} & \parbox{7mm \centering \textsc{fit+sd}} & \parbox{6mm \centering\textsc{fit+s}} & 
& \textsc{seed} & \parbox{6mm \centering \textsc{fit}}  & \parbox{1cm \centering\textsc{fit+sw}} & \parbox{7mm \centering \textsc{fit+sd}}
      & \parbox{6mm \centering\textsc{fit+s}} & \\\cline{1-7}\cline{9-13}
      \multirow{2}{*}{GLoVE}  & r$^+$-acc &
                       \parbox{5.8mm}{\centering 0.74} & \parbox{6mm}{\centering 0.59} & \parbox{1cm}{\centering 0.57} & \parbox{6.4mm}{\centering \textbf{0.76}} & \parbox{6.4mm}{\centering 0.72} & \multirow{2}{*}{\parbox{9mm}{\textsc{freq} r$^+$-acc: 0.65 \vspace{1mm} \\ {\sc rand} r$^+$-acc: 0.50}}  &  
                       \parbox{5.8mm}{\centering \textbf{0.73}} & \parbox{6mm}{\centering 0.53} & \parbox{6.7mm}{\centering 0.37} & \parbox{6.4mm}{\centering 0.68} & \parbox{6.4mm}{\centering 0.69} & \multirow{5}{*}{\parbox{9mm}{{\sc freq} r$^+$-acc: 0.63  
                       \vspace{1mm} \\ {\sc rand} r$^+$-acc: 0.51}} 
                       \\ 
      & MSE &
              \parbox{5.8mm}{\centering 31.5} & \parbox{6mm}{\centering 24.3} & \parbox{1cm}{\centering 75.1} & \parbox{6.4mm}{\centering 1.5} & \parbox{6.4mm}{\centering \textbf{1.2}} & & 
              \parbox{5.8mm}{\centering 60.5} & \parbox{6mm}{\centering 396.5} & \parbox{6.7mm}{\centering 285.7} & \parbox{6.4mm}{\centering 1.8} & \parbox{6.4mm}{\centering \textbf{1.6}} &  \\ \cline{1-7}\cline{9-13}
      \multirow{2}{*}{BERT} & r$^+$-acc &
                      \parbox{5.8mm}{\centering 0.69} & \parbox{6mm}{\centering 0.52} & \parbox{1cm}{\centering 0.52} & \parbox{6.6mm}{\centering 0.71} & \parbox{6.6mm}{\centering \textbf{ 0.72}} &  &
                       \parbox{5.8mm}{\centering 0.64} & \parbox{6mm}{\centering 0.52} & \parbox{6.7mm}{\centering 0.51} & \parbox{6.4mm}{\centering 0.64} & \parbox{6.4mm}{\centering \textbf{0.69}} &  \\                                          
      & MSE &
              \parbox{5.8mm}{\centering 123.5} & \parbox{6mm}{\centering 437.2} & \parbox{1cm}{\centering 724.0} & \parbox{6.4mm}{\centering 3.6} & \parbox{6.4mm}{\centering \textbf{2.4}} & &  
             \parbox{5.8mm}{\centering 215.6} & \parbox{6mm}{\centering 216.3} & \parbox{6.7mm}{\centering 617.1} & \parbox{6.4mm}{\centering 7.8} & \parbox{6.4mm}{\centering \textbf{3.2}} &  \\ \cline{1-7}\cline{9-13}                                            
      \multirow{2}{*}{RoBERTa}  & r$^+$-acc &
                         \parbox{5.8mm}{\centering \textbf{0.74}} & \parbox{6mm}{\centering 0.51} & \parbox{1cm}{\centering 0.51} & \parbox{6.4mm}{\centering 0.71} & \parbox{6.4mm}{\centering 0.73} & & 
                         \parbox{5.8mm}{\centering 0.67} & \parbox{6mm}{\centering 0.53} & \parbox{6.7mm}{\centering 0.53} & \parbox{6.4mm}{\centering 0.66} & \parbox{6.4mm}{\centering \textbf{0.71}} &  \\                         
      & MSE &
              \parbox{5.8mm}{\centering 82.7} & \parbox{6mm}{\centering 591.9} & \parbox{1cm}{\centering $>1000$} & \parbox{6.4mm}{\centering 3.9} & \parbox{6.4mm}{\centering \textbf{2.3}}& & 
                                                            \parbox{5.8mm}{\centering 223.0} & \parbox{6mm}{\centering 325.0} & \parbox{6.7mm}{\centering 778.1} & \parbox{6.7mm}{\centering 7.3} &  \parbox{6.4mm}{\centering \textbf{2.4}} &  \\  
    \end{tabular}
    \caption{Results on {\bf formality} and {\bf complexity}: Extended rank
      accuracy (r$^+$-acc) 
      and Mean Squared Error (MSE). Shown for 3 embedding spaces. Bolded: best performance for each embedding.}
    \label{tab:formalitycomplexity}
  \end{table*}

 \paragraph{Overall performance.}

Overall results are shown for object properties in
Table~\ref{tab:grand} and for stylistic features in
Table~\ref{tab:formalitycomplexity}. 
Looking at object properties first, and focusing on extended rank accuracy, the \textsc{fit} model by
itself is not very good, and adding seeds as words (\textsc{fit+sw})
does not help. \textsc{fit+sd} is better, and outperforms
\textsc{seed} slightly, but the \textsc{fit+s}
model, which computes fitted dimensions using seed information both as
words and dimensions, shows the best performance, outperforming
\textsc{seed} strongly. In terms of MSE, even medians are very
high for the \textsc{seed} model, so many 
seed-based dimensions were not able to predict ratings on the same scale as the gold
ratings. MSE is much lower for both fitted models that make use of
seed dimensions, especially \textsc{fit+s}.\footnote{Though
  note that the ratings are z-scores, so the MSE is still larger than
  half a standard deviation.} Looking at the baselines, the
\textsc{fit} and \textsc{fit+sw} models with BERT and RoBERTa
underperform the frequency baseline, and are on par with random
guessing. The frequency baseline is somewhat higher than random, though it is not entirely clear what kind of signal for object properties can be derived from word frequency. 

On the stylistic data, the relative performance of the fitted models 
is similar, but here they mostly do not outperform the \textsc{seed}
dimensions. Overall performance of the \textsc{seed} dimensions is
higher here, which raises the question if fitted models help in
particular when seed-based dimensions do not perform well; we explore
this further below. Looking at MSE, we confirm that fitted models that
use seed dimensions have much lower error than the other
models. 
Comparing embedding spaces, we see consistently the best performance
with GLoVE embeddings. The BERT and RoBERTa \textsc{fit} and
\textsc{fit+sw} models in particular are again at the level of the
random baseline. The frequency baseline is reasonably strong, matching
previous findings.

\paragraph{Fitted dimensions, by themselves, are underdetermined by
  human ratings.} The \textsc{fit} model, which computes fitted
dimensions from human ratings only, does not perform well, and we
suspect that the size of the embedding space allows for too many ways
to fit a dimension to ratings, causing the model to overfit. To test
this, we first computed \textsc{fit} dimensions, for Grand et al.\
object features, from \emph{all} human ratings, obtaining perfectly
fit dimensions in every single case. We next train dimensions on all human ratings
but scramble the word/rating pairs, making them nonsensical. Again we
obtain perfectly fit dimensions in every single case, which confirms
our suspicion. The picture that emerges is that \textsc{fit} by itself
does not have enough information to fit a good dimension and overfits
to the training data. The seed information provided to 
\textsc{fit+sw}, \textsc{fit+sd} and \textsc{fit+s} gives the models
the additional guidance needed to make good use of the human ratings, and
the combination of seeds and human ratings on words leads to overall
better dimensions -- at least in some cases. We next ask which cases
those are.

\begin{figure}[tb]
  \centering
  \includegraphics[width=14em]{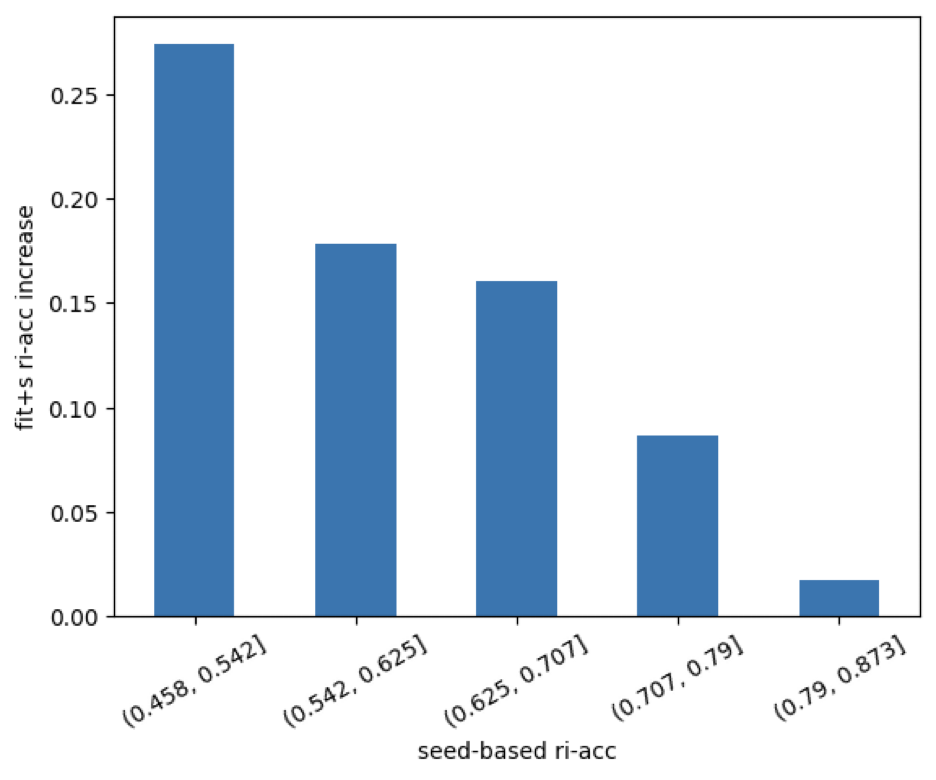}
  \caption{Increase in \textit{r$^+$-acc} for 
    \textsc{fit+s} over \textsc{seed}, object properties grouped by performance of \textsc{seed}.}
  \label{fig:numimproved}
\end{figure}

\paragraph{Human ratings help most when seed-based dimensions underperform.} Comparing \textit{r$^+$-acc}
values for seed-based dimensions and \textsc{fit+s} dimensions on the
object property data, we find that
\textsc{fit+s} improves over \textsc{seed} in every single one of the 50
category/property pairs.\footnote{The analysis includes all conditions not in the
  development set. For each condition, we again ran 5-fold cross-validation,
  each repeated over 3 random seeds, then averaged by condition.}
The performance increase is
highest when performance of the seed-based dimensions is lowest, as
shown in Figure\ \ref{fig:numimproved}: For the 20\% of
category/property pairs with lowest
\textsc{seed} performance, average improvement is 27.3
points, while for the 20\% of category/property pairs with the highest \textsc{seed}
performance, average improvement is 1.7 points. This could explain
the lack of improvement achieved on stylistic features, as
\textsc{seed} already performs well on this data.

Table~\ref{tab:granddetail} further zooms in on the object feature
data, showing performance on some category/property pairs with low, medium, and
high performance of the \textsc{seed} dimensions. We see that
\textsc{fit} and \textsc{fit+sw}  underperform
throughout. \textsc{fit+s} shows the overall best performance, but the
improvement over \textsc{seed} is particularly high for the first
group of conditions, where \textsc{seed} dimensions get no traction on
the data. \textsc{fit+sd} shows good extended rank accuracy on the conditions with
medium to high \textsc{seed} performance, but not on the conditions
that are particularly poorly modeled by \textsc{seed}. 

\begin{table}[tb]
  \centering
  \small
  \scalebox{0.85}{
  \begin{tabular}{l|*{5}{r}|*{5}{l}}
 & \multicolumn{5}{c}{\bf r$^+$-acc} \\ 
 \textbf{Category, Feature} & {\sc seed} & {\sc fit} & {\sc fit+sw} &
{\sc fit+sd} & {\sc fit+s} \\\hline
sports/speed & 0.46 &  0.55 &  0.56 &  0.52 &  0.78 \\
states/cost & 0.46 &  0.5 &  0.5 &  0.42 &  0.83 \\
cities/arousal & 0.47 &  0.52 &  0.5 &  0.51 &  0.82 \\
animals/intelligence & 0.48 &  0.55 &  0.54 &  0.5 &  0.79 \\
clothing/cost & 0.48 &  0.52 &  0.51 &  0.55 &  0.76 \\\hline

 clothing/wealth & 0.62 &  0.52 &  0.53 &  0.6 &  0.82 \\
states/wealth & 0.62 &  0.55 &  0.56 &  0.64 &  0.82 \\
 weather/temperature & 0.69 &  0.56 &  0.52 &  0.66 &  0.76 \\
animals/danger & 0.7 &  0.6 &  0.57 &  0.76 &  0.84 \\
 clothing/age & 0.71 &  0.52 &  0.55 &  0.71 &  0.8 \\\hline

weather/danger & 0.79 &  0.55 &  0.54 &  0.7 &  0.82 \\
clothing/gender & 0.81 &  0.56 &  0.53 &  0.8 &  0.82 \\
sports/gender & 0.81 &  0.61 &  0.56 &  0.81 &  0.84 \\
professions/gender & 0.85 &  0.56 &  0.56 &  0.87 &  0.86 \\
names/gender & 0.87 &  0.56 &  0.51 &  0.87 &  0.87 \\
\end{tabular}}
  \caption{Detail results for Grand et al. by \textsc{seed}
    performance: lowest performance (top box), middling performance
    (middle), best performance (bottom).}
\label{tab:granddetail}
\end{table}

\begin{figure}[tb]
  \centering
  \includegraphics[width=14em]{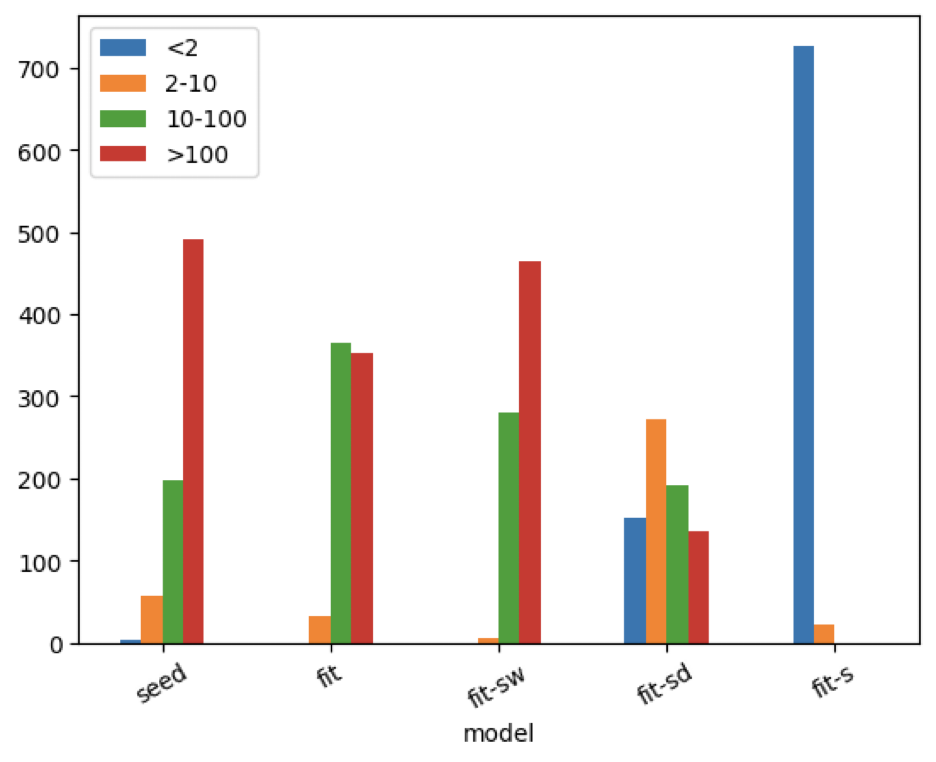}
  \caption{MSE distributions for runs of different models. y-axis:
    count of runs.}
  \label{fig:mse}
\end{figure}

\paragraph{\textsc{fit+s} models are the only ones that predict
  ratings on the gold scale.} We saw above that
median MSE values are extremely high for many models, especially for
\textsc{seed}. We now take a closer look, in particular we want to
know how often we obtain MSE values that are extremely far off from the gold ratings. We
again focus on the object feature data as we there have many
conditions that we can compare. Figure\
\ref{fig:mse} shows, for each model, how many runs had MSE values
in the ranges of $<2, 2-10, 10-100$, and $>100$. Recall that gold
ratings are z-scores, so they tend to be in a range of -2 to 2. We again only use the
category/property pairs that are not in the development set, but now count
separately each cross-validation run and each random seed. We see that
many runs of \textsc{seed}, \textsc{fit} and \textsc{fit+sw} have very
high MSE values. In \textsc{fit+sd} we first see a considerable
percentage of runs with MSE values
below 2 (the blue bar comprises 20\% of runs for this model), but
strikingly, 97\% runs of \textsc{fit+s} have
MSE values below 2, and all have values below 10. So this model is much more consistent than the
other models, and in fact is highly consistent in fitting
dimensions that deliver predictions in the range of the gold data.

\paragraph{Zooming in: Examples of predictions.}

\begin{figure}[tb]
  \centering
  \begin{tabular}{@{}ll}
    \includegraphics[width=9.5em]{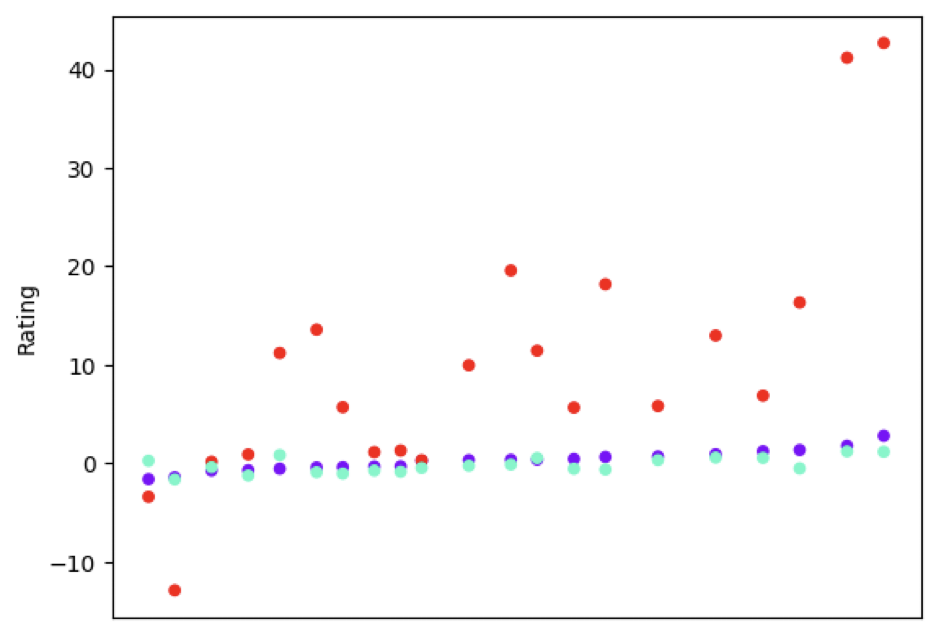} &
   \includegraphics[width=9.5em]{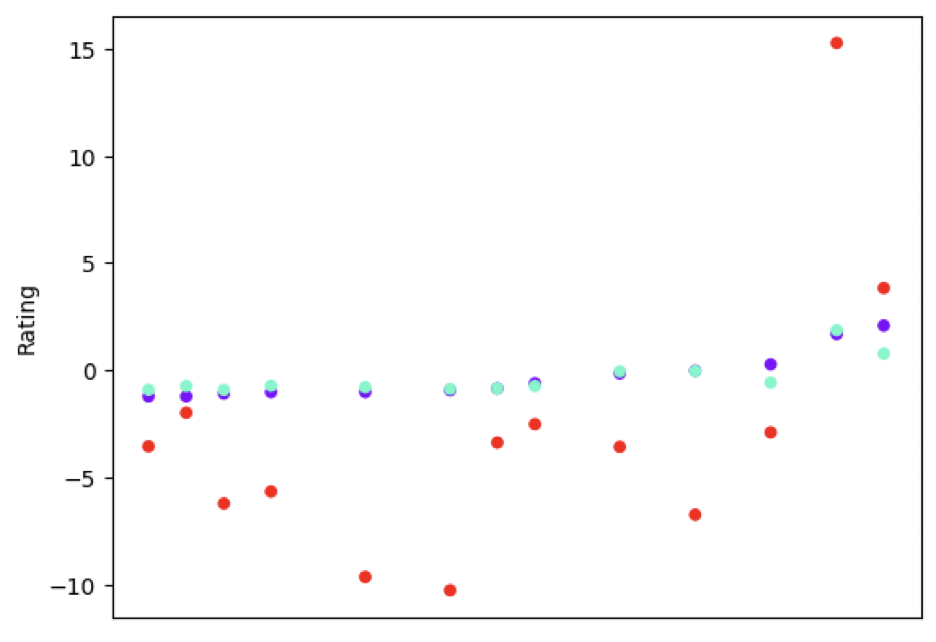}                                                     
  \end{tabular}
  \caption{Clothes rated for  wealth (left), animals rated for size
    (right). Gold ratings: dark purple. \textsc{seed} predictions:
    red. \textsc{fit+s} prediction:  light blue. Datapoints ordered by
  gold rank.}
  \label{fig:errplot}
\end{figure}

\begin{table}[tb]
  \centering
  \small
  \begin{tabular}{lll}
    Gold & \textsc{seed} & \textsc{fit+s} \\\hline
bee & chipmunk & butterfly \\
ant & hamster & bee \\
butterfly & \textit{monkey} & \textit{chipmunk} \\
goldfish & butterfly & bird \\
hamster & goldfish & hamster \\
chipmunk & dog & \textit{ant} \\
bird & \textit{bee} & crow \\
crow & bird & \textit{goldfish} \\
dog & seal & seal \\
monkey & crow & dog \\
seal & \textit{ant} & monkey \\
mammoth & whale & whale \\
    whale & mammoth & mammoth \\
  \end{tabular}
  \caption{Comparing word rankings by humans, \textsc{seed}
    dimensions, and \textsc{fit+s} dimensions: Animals by
    size. Italicized: 3 words with highest error in ranking.}
  \label{tab:rankingexample}
\end{table}

\begin{table}[tb]
  \centering
  \small
  \begin{tabular}{lll}
    Gold & \textsc{seed} & \textsc{fit+s} \\\hline
sweatshirt & shorts & shorts \\
shorts & sweatshirt & boots \\
belt & belt & bikini \\
boots & blouse & tights \\
hat & boots & skirt \\
tights & swimsuit & swimsuit \\
bikini & skirt & stockings \\
swimsuit & trousers & trousers \\
skirt & bikini & \textit{loafers} \\
blouse & robe & blouse \\
knickers & cuff & belt \\
dress & knickers & knickers \\
collar & \textit{hat} & dress \\
trousers & collar & \textit{sweatshirt} \\
stockings & vest & robe \\
robe & \textit{tights} & collar \\
vest & loafers & cuff \\
cuff & stockings & vest \\
loafers & \textit{dress} & \textit{hat} \\
gown & gown & tuxedo \\
    tuxedo & tuxedo & gown \\
  \end{tabular}  
  \caption{Comparing word rankings by humans, \textsc{seed}
    dimensions, and \textsc{fit+s} dimensions: Clothes by wealth. Italicized: 3 words with highest error in ranking.}
  \label{tab:rankingexamplewealth}
\end{table}

We take a closer look at two kinds of object properties: clothes by
wealth, and animals by size. In order to obtain sufficiently many test
datapoints to look at, we divide the data into 2/3 training and 1/3
test (as opposed to the 1/5 we use with 5-fold
cross-validation). Figure~\ref{fig:dims} shows the test data words,
along with seed-based and \textsc{fit+s} dimensions, downprojected
into 2 dimensions using PCA. For the same datapoints,
Figure~\ref{fig:errplot} plots gold ratings, \textsc{seed}
predictions, and \textsc{fit+s} predictions. This plot illustrates how
the \textsc{seed} predictions are on a much larger scale than gold
ratings, while \textsc{fit+s} is the only model whose predictions stay
on the same scale. (The next to last datapoint in animals/size is
\textit{mammoth}, which \textit{seed} largely overestimates -- maybe
because \textit{mammoth} is also an adjective indicating gargantuan
size.) Tables~\ref{tab:rankingexample} and
\ref{tab:rankingexamplewealth} show how the test data words
for animals by size, and for clothes by wealth, are ranked by humans, by the \textsc{seed} dimension,
and by the \textsc{fit+s} dimension. The italicized words are the three words
whose model rank is furthest off from their gold rank. For the animals
data, both models mis-rank \textit{ant}, and overall seem to struggle
more with smaller animals. Among the clothes, both models overestimate
the wealth projected by wearing hats.

\section{Conclusion}

In this paper we have  proposed a method for constructing high quality interpretable dimensions in embedding spaces. 
We show that by combining seed-based vectors with guidance from human ratings about  properties, it is possible to induce better  dimensions than with the seed-based methodology alone. We expect the proposed dimensions to be useful in various areas of study, including linguistics, psychology, and social science. 

For the moment, the  proposed dimensions address one property at a time. In future work, we are planning to explore multifaceted properties which would be better represented through multiple dimensions. Aside from a more elaborate description of these properties, a  space of multiple interpretable dimensions will offer a rich context of comparison for words that might be similar in some respect and not in others (e.g., \textit{tiger} and \textit{spider} with respect to {\sc danger} and {\sc size}).

\section{Limitations}

In our experiments we use English models and data. The seed-based methodology has been shown to work well in other languages, so an extension of the proposed methodology to other languages is possible. A limitation regarding this extension is the lack of human ratings which are needed for calculating the fitted dimensions. A possible mitigation would be to translate the annotated English data into other languages. 

The ratings we used in our study were averages over individual human ratings, possibly obscuring legitimate differences between raters \citep{plank-etal-2014-linguistically}. Another limitation of the human ratings used in this study is that they were out of context, possibly obscuring effects of topic and polysemy. 

There are many different ways to use contextualized embeddings. We have averaged over all token representations generated by BERT and RoBERTa for a word in a sentence pool, and used the top 4 layers of the models. 
It is possible that BERT and RoBERTa would do better, or at least differently, if other model layers (or layer combinations) were used.

Our approach is not at all compute intensive. All computations were done on a laptop.

\section*{Acknowledgements}

We would like to thank the anonymous reviewers for their valuable feedback. This research is supported in part by the Office of the Director of National Intelligence (ODNI), Intelligence Advanced Research Projects Activity (IARPA), via the HIATUS Program contract \#2022-22072200005. The views and conclusions contained herein are those of the authors and should not be interpreted as necessarily representing the official policies, either expressed or implied, of ODNI, IARPA, or the U.S. Government. The U.S. Government is authorized to reproduce and distribute reprints for governmental purposes notwithstanding any copyright annotation therein. 

\bibliography{anthology,custom}

\appendix

\section{Appendix}
\label{sec:appendix}
\paragraph{Details on computing.} All experiments were conducted on a
MacBook Pro laptop using Python 3.8, with huggingface version 4.35.2, torch version 2.0.1, sklearn
version 1.2.1, numpy version 1.22.4 and scipy 1.10.0. 

\paragraph{Hyperparameter estimation.} Development set: 6 conditions
sampled at random from the object features dataset: cities-danger, states-political, 
  animals-wetness, cities-intelligence,
  animals-weight, names-age.

  As said above, only hyperparameters that made a difference:
  averaging, always good, and mixing parameter alpha. Chosen values: 

Best parameters:
\begin{itemize}
\item $\alpha$ for \textsc{fit+sd}: GLoVE 0.02
\item $\alpha$ for \textsc{fit+s}: GLoVE 0.05
\end{itemize}

\paragraph{Embedding spaces, and sentence selection.} 

The GLoVe embeddings used were 
trained on Common Crawl (42B tokens, 1.9M
vocab, uncased, 300d vectors), downloaded from \url{https://nlp.stanford.edu/projects/glove/}.

In order to generate embeddings for contextualized instances of words in our datasets using BERT ({\tt bert-large-uncased}) and RoBERTa ({\tt roberta-large}) 
models \citep{devlin2019bert,liu2019roberta}, we used sentences from the UkWac corpus \citep{baroni2009wacky}. We collected ten sentences for each word, when available. We filtered out sentences with more than 100 tokens in order to avoid including noisy contexts such as webpage menus crawled from the web.  
If a word had less than 10 occurrences in UkWac, we used as many sentences as were available. This was the case for 10 words in the Grand et al. dataset (\textit{nairobi} (6), \textit{seoul} (4), \textit{taipei} (5), \textit{lahore} (2), \textit{baghdad} (9), \textit{peyton} (9), \textit{tehran} (4), \textit{johannesburg} (4), \textit{jaime} (5), \textit{karachi} (7); and for one word (\textit{jazeera} (6)) in the formality dataset. For hyphenated words in the Grand et al. dataset (e.g., \textit{new-york, south-carolina, south-dakota}), we collected sentences where they occur without the hyphen.

\end{document}